\documentclass[letterpaper, 10 pt, conference, twoside]{IEEEconf}

\IEEEoverridecommandlockouts                              

\usepackage{amsmath,amssymb,amsfonts}
\usepackage{calrsfs}
\DeclareMathAlphabet{\pazocal}{OMS}{zplm}{m}{n}
\usepackage{cite}
\usepackage{graphicx}
\usepackage{mathrsfs}
\usepackage{xcolor}
\usepackage{balance}

\newtheorem{theorem}{\textbf{Theorem}}

\newtheorem{definition}{Definition}

\newcommand{\Int}{\mathbb{Z}_{\geq 0}}
\newcommand{\Real}{\mathbb{R}}

\newcommand{\R}{\mathbb{R}}

\newcommand{\col}{\textrm{col}}

\newcommand{\ini}{\textrm{ini}}
\newcommand{\des}{\textrm{des}}

\title{\LARGE \bf Distributed Data-Driven Predictive Control for Multi-Agent Collaborative Legged Locomotion}
\author{Randall~T.~Fawcett$^{1}$, Leila~Amanzadeh$^{1}$, Jeeseop Kim$^{1}$, Aaron~D.~Ames$^{2}$, and Kaveh~Akbari~Hamed$^{1}$
\thanks{The work of R. T. Fawcett is supported by the National Science Foundation (NSF) under Grant 2128948. The work of K. Akbari Hamed is supported by the NSF under Grants 1924617 and 2128948. The work of A. D. Ames is supported by the NSF under Grant 1924526.}
\thanks{$^{1}$R. T. Fawcett, L. Amanzadeh, J. Kim, and K. Akbari Hamed (Corresponding Author) are with the Department of Mechanical Engineering, Virginia Tech, Blacksburg, VA, 24061, USA, {\tt\small \{randallf, leila7, jeeseop, kavehakbarihamed,\}@vt.edu}}%
\thanks{$^{2}$A. D. Ames is with the Department
of Mechanical and Civil Engineering, California Institute of Technology, Pasadena, CA, 91125, USA, {\tt\small ames@caltech.edu}}%
}

\begin{document}

\maketitle
\pagestyle{empty} 
\thispagestyle{empty} 

\begin{abstract}
    The aim of this work is to define a planner that enables robust legged locomotion for complex multi-agent systems consisting of several holonomically constrained quadrupeds. To this end, we employ a methodology based on behavioral systems theory to model the sophisticated and high-dimensional structure induced by the holonomic constraints. The resulting model is then used in tandem with distributed control techniques such that the computational burden is shared across agents while the coupling between agents is preserved. Finally, this distributed model is framed in the context of a predictive controller, resulting in a robustly stable method for trajectory planning. This methodology is tested in simulation with up to five agents and is further experimentally validated on three A1 quadrupedal robots subject to various uncertainties including payloads, rough terrain, and push disturbances.  
\end{abstract}


\begin{keywords}
Legged Robots, Motion Control, Multi-Contact Whole-Body Motion Planning and Control
\end{keywords}
\vspace{-0.1em}

\vspace{-0.5em}
\section{Introduction}\label{INTRODUCTION}
\vspace{-0.5em}

\begin{figure}[t!]
\centering
\includegraphics[width=0.75\linewidth]{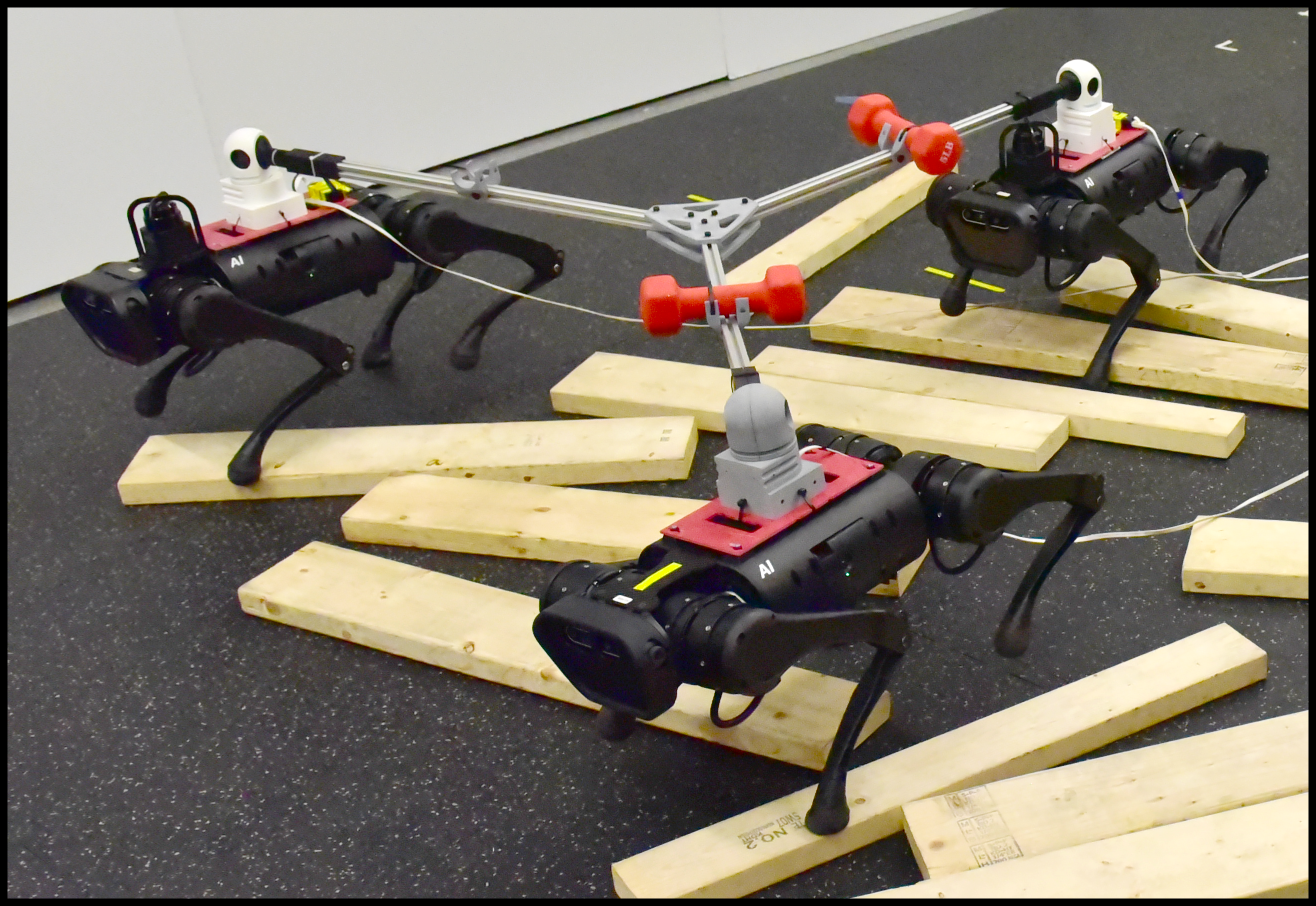}
\vspace{-1.em}
\caption{Snapshot showing the locomotion of three holonomically constrained A1 robots on wooden blocks.}
\label{MainSnap}
\vspace{-1.5em}
\end{figure}

This work investigates multi-agent systems composed of high-dimensional quadrupedal robots that are rigidly holonomically constrained to one another using ball joints, introducing high interaction forces (see Fig. \ref{MainSnap}). Current state-of-the-art approaches, even when considering only a single agent, generally involve using a reduced-order model of some kind \cite{SLIP_02}. For a single agent, there have been many template models that have worked effectively as will be discussed shortly. However, template models for multi-agent systems, particularly those with large interaction forces, have not yet been fully explored. In particular, it is difficult to use traditional modeling techniques to model an increasing number of agents due to increased dynamic complexity. Even in the case that a large-scale dynamical system with strong interaction forces could be modeled at the reduced-order level, it would likely be of such complexity that it would no longer \textit{efficiently} function as a reduced-order model for control. The goal of this work, therefore, is to address these issues. Namely, we aim to synthesize data-driven template models for planning of large multi-agent systems such that the resulting planner is computationally efficient for use in real-time.

\vspace{-0.3em}
\subsection{Multi-Agent Systems}
\vspace{-0.2em}

Multi-Agent systems have been a major area of research, particularly in the context of cooperative manipulation \cite{culbertson2021decentralized,Murray_Book}, control of unmanned aerial vehicles \cite{chen2020guaranteed,RSS2013}, and ground vehicles \cite{DP:MT:14,machado2016multi}. However, when considering many agents, the system is usually high-dimensional, which also motivates distributed approaches \cite{DUNBAR_distributedMPC, Bullo_Book, Mesbahi_Book}. These methods reduce the computational burden for controllers and planners, but they have also \textit{not} been readily extended to legged robotic systems with high dimensionality, unilateral constraints, and high interaction forces, as is the case in this work. Planning for multi-agent systems also usually necessitates the use of a reduced-order model \cite{perizzato2015formation}, but in the case of holonomically constrained quadrupeds, the best model is unclear.

\begin{figure*}[t!]
\centering
\vspace{0.1em}
\includegraphics[width=\linewidth]{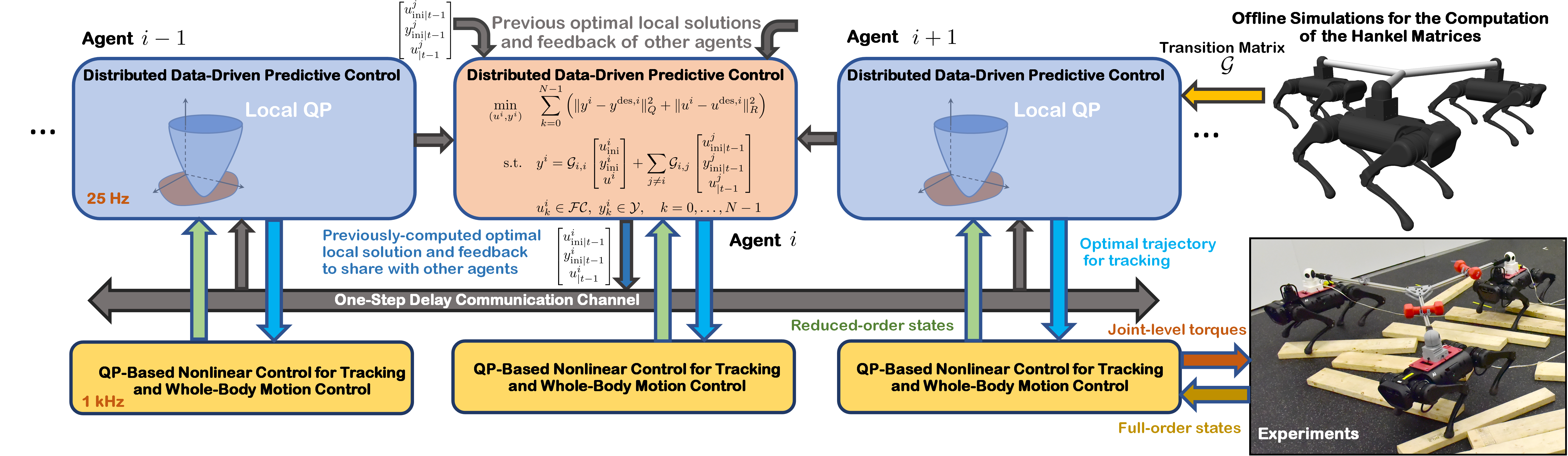}
\caption{Overview of the proposed control algorithm with distributed DDPC algorithms at the high level for trajectory optimization of cooperative locomotion and nonlinear controllers at the low level for tracking and whole-body motion control.}
\label{OverviewFig}
\vspace{-1.8em}
\end{figure*}

\vspace{-0.3em}
\subsection{Reduced-Order Models}
\vspace{-0.3em}

In order to create predictive controllers for complex systems, it is often necessary to utilize a reduced-order model. In terms of legged locomotion, a very well-known and studied model is that of the linear inverted pendulum (LIP) \cite{LIP_Original_Kajita}, which has been used for both quadrupedal and bipedal locomotion. However, the LIP model generally enforces quasi-static motions due to the requirement that the center of pressure remains within the convex hull formed by the contacting points with the environment. Nonetheless, the model's simplicity has been of great interest over the years and has been used extensively in both simulation and experiments using various platforms \cite{Leonessa_Pratt_MPC,Pratt_LIP,Ott_MPC,LIP_Original_Kajita,Hamed_Kim_Pandala_MPC}. It is also worth mentioning that this model is not amenable to adding moments induced about the center of mass (COM) by external forces \cite{kim2022cooperative,Randy_Hamed_ASME}, which is important when dealing with holonomically constrained multi-agent systems. 


More recently, the single rigid body (SRB) model has gained traction to achieve dynamic locomotion \cite{KF_mitcheetah3}. This reduced-order model has proven to be effective in creating natural and dynamic motions for various legged robot platforms \cite{Ames_Hutter_ICRA_MPC_CBFs,Wensing_VBL_HJB, Abhishek_Hae-Won_TRO,KF_mitcheetah3}. However, a drawback is that the legs are assumed to be massless for this model, which can become difficult when using robots with more mass. Conversely, one major benefit of the SRB model, in contrast to the LIP model, is that it allows one to consider the moments induced about the COM caused by external forces, which is useful in this case because of the holonomic constraints. Unfortunately, using the SRB model subject to interaction forces among holonomically constrained agents becomes prohibitively complex as the number of agents increases, which motivates the use of data-driven methods.

\vspace{-0.3em}
\subsection{Data-Driven Methods}
\vspace{-0.3em}

Data-driven methods are becoming more popular as systems become more complex \cite{hou2013model}. This motivates their use in multi-agent systems, especially in the case where the individual agents are complex. Even though individual systems can be modeled adequately, their combined dynamics make the system increasingly sophisticated, and an effective reduced-order model has yet to be determined. One such method recently popularized in the robotics community is reinforcement learning \cite{Leeeabc5986,miki2022learning}. This method has been used for collaborative locomotion among multiple quadrupeds \cite{ji2021reinforcement} but takes considerable time and computation power. Furthermore, reinforcement learning provides little intuition as to the inner workings of the result. In this work, we pursue a different avenue.

Here, we consider the use of data-driven methods in the context of behavioral systems theory. This methodology parameterizes a linear time-invariant (LTI) system  directly in terms of its measured trajectories \cite{Willems_FundamentalThm,willems1986time,markovsky2006BST}. Behavioral systems theory, when used in a predictive control or planning framework, is usually referred to as data-driven predictive control (DDPC) or data-enabled predictive control (DeePC) and has lately become of great interest to the robotics community \cite{Berberich_DDPC,ShallowsDeePC,PowerSysDeePC,coulson2021distributionally,Waarde_DDPC,wei2020data,Fawcett_Hamed_DataDriven}. Recently, extensions have been made to use such methods for certain types of nonlinear systems \cite{berberich2020trajectory} and linear parameter varying systems \cite{verhoek2021LPV}, but there have yet to be any proofs extending to a broad class of general nonlinear systems. There have also been advances in using these methods for stochastic implementations, and in particular, there have been experimental validations for nonlinear systems, even though the theory does not directly apply \cite{ShallowsDeePC,PowerSysDeePC,coulson2021distributionally,Fawcett_Hamed_DataDriven}. However, to the best of the authors' knowledge, there have not been implementations for multi-agent systems. We further note that the extension to collaborative legged locomotion introduces many difficulties, including the hybrid nature of legged locomotion and unilateral constraints.

\vspace{-0.3em}
\subsection{Goals, Objectives, and Contributions}
\vspace{-0.3em}

The \textit{overarching goal} of this paper is to develop a computationally tractable real-time predictive planner for collaborative legged locomotion using a behavioral approach. Namely, this work's \textit{objectives} and \textit{key contributions} are enumerated as follows: 1) A model is created using concepts from behavioral systems theory for systems of quadrupeds that are holonomically constrained to one another. 2) The model is used in the context of a \textit{distributed} predictive control framework such that each agent can effectively plan for its own motions while considering the motions of other agents (see Fig. \ref{OverviewFig}). 3) Simulation results for $5$ constrained quadrupeds in the presence of uncertainty are provided. 4) We present extensive experimental validation on a team of $3$ holonomically constrained quadrupeds. The experimental validation shows robust locomotion of the A1 robot subject to various uncertainties, including rough terrain, push disturbances, and outdoor environments.


\vspace{-0.5em}
\section{Preliminaries}
\label{PRELIMINARIES}
\vspace{-0.3em}

In this section, we overview some important concepts from behavioral systems theory that will be used throughout this work. Behavioral systems theory provides a manner in which data collected from a system can be leveraged to directly create a model. In particular, consider an LTI model with the state vector $x_{k}\in\R^{n}$, the input vector $u_{k}\in\R^{m}$, and the output vector $y_{k}\in\R^{p}$ for $k\in\Int:=\{0,1,\cdots\}$. Such a model can be represented in discrete time as follows:
\begin{alignat}{1}\label{LTI System}
    x_{k+1} &= A\,x_{k} + B\,u_k \nonumber \\
    y_{k} &= C\,x_{k} + D\,u_k,
\end{alignat}
where $A\in\R^{n\times n}$, $B\in\R^{n\times m}$, $C\in\R^{p\times n}$, and $D\in\R^{p\times m}$ are the state-space matrices, which are unknown. In this notation, $n$, $m$, and $p$ represent the number of states, inputs, and outputs, respectively. One difference between traditional system identification and behavioral systems theory is that, traditionally, one would attempt to reconstruct the matrices of $\eqref{LTI System}$ using data. In the behavioral context, we obtain an input-output (I-O) model without reconstructing the state matrices. 

To introduce the concepts, consider $L,T\in\Int$, where $T\geq L$. In addition, define some input trajectory $u^{\textrm{d}}\in\R^{mT}$ composed of a sequence of $u^\textrm{d}_{k}$, i.e., $u^\textrm{d}:=\col(u^\textrm{d}_{0},\hdots,u^\textrm{d}_{T-1})$. In this notation, ``$\textrm{col}$'' represents the column operator. Using this trajectory, one can construct the following Hankel matrix
\begin{equation}\label{Hankel}
    \mathcal{H}_{L}(u^\textrm{d}) := \begin{bmatrix}
        u^\textrm{d}_{0} & u^\textrm{d}_{1} & \cdots & u^\textrm{d}_{T-L} \\
        u^\textrm{d}_{1} & u^\textrm{d}_{2} & \cdots & u^\textrm{d}_{T-L+1} \\
        \vdots & \vdots & \ddots & \vdots \\
        u^\textrm{d}_{L-1} & u^\textrm{d}_{L} & \cdots & u^\textrm{d}_{T-1}
    \end{bmatrix}\in\Real^{mL\times{(T-L+1)}}.
\end{equation}
\begin{definition}[\cite{Berberich_DDPC}]\label{HankelDef}
    The signal $u^{\textrm{d}}$ is said to be \textit{persistently exciting} of order $L$ if $\mathcal{H}_{L}(u^{\textrm{d}})$ is full row rank.
\end{definition}

\begin{definition}[\cite{Berberich_DDPC}]\label{TrajectoryDef}
The sequence $\{(u_{k},y_{k})\}_{k=0}^{T-1}$ is said to be a \textit{trajectory} of the LTI system \eqref{LTI System} if there exists an initial condition $x_{0}$ and a state sequence $\{x_{k}\}_{k=0}^{T}$ that meets the state and output equations in \eqref{LTI System}.
\end{definition}

Using Definitions \ref{HankelDef} and \ref{TrajectoryDef}, we can now present a foundational theorem in behavioral systems theory, used to represent an LTI system based on observed trajectories.

\begin{theorem}{\cite[Theorem 1]{Willems_FundamentalThm}}\label{BST_Theorem}
    \textit{Let an observed trajectory of \eqref{LTI System}, referred to as data, be denoted by $\{(u^\textrm{d}_k,y^\textrm{d}_k)\}_{k=0}^{T-1}$. If $u^\textrm{d}$ is persistently exciting of order $L+n$, then $\{(\bar{u}_k,\bar{y}_k)\}_{k=0}^{L-1}$ is a trajectory of the system if and only if there exists $g\in\Real^{T-L+1}$ such that}
    \begin{equation}\label{FundamentalLemma}
        \begin{bmatrix} \mathcal{H}_{L}(u^\textrm{d}) \\ \mathcal{H}_{L}(y^\textrm{d}) \end{bmatrix}g = \begin{bmatrix} \bar{u} \\ \bar{y} \end{bmatrix}.
    \end{equation}
\end{theorem}
\vspace{0.5em}

Theorem \ref{BST_Theorem} provides a constructive manner in which an LTI system can be represented through its trajectories without system identification. This concept will be used throughout this work as we aim to parameterize a complex system of robots by using only their trajectories. To do so, we consider two different horizons denoted by $T_{\ini}$ and $N$, which represent the estimation horizon and control horizon, respectively. The estimation horizon encapsulates the input-output pairs that are required in order to determine the initial conditions of a trajectory given a particular I-O sequence $\{(\bar{u}_{k},\bar{y}_{k})\}_{k=0}^{L-1}$ from \eqref{FundamentalLemma}. In contrast, the prediction horizon represents how far into the future predictions of the trajectories are made, similar to that of traditional model predictive control (MPC). Finally, we define $L=T_{\ini}+N$ for compact notation. We denote the collected I-O data by $(u^\textrm{d},y^\textrm{d})$ and can decompose the Hankel matrices of \eqref{FundamentalLemma} into two parts as follows:
\begin{alignat}{2}\label{HankelDecomp}
    \mathcal{H}_{L}(u^\textrm{d}) = \begin{bmatrix} U_{p} \\ U_{f} \end{bmatrix}, \quad & \mathcal{H}_{L}(y^\textrm{d}) = \begin{bmatrix} Y_{p} \\ Y_{f} \end{bmatrix},
\end{alignat}
where $U_p\in\R^{mT_\ini \times (T-L+1)}$ and $Y_p\in\R^{pT_\ini \times (T-L+1)}$ are the portions of the Hankel matrices used for estimating the initial condition (i.e., past), and $U_f\in\R^{mN \times (T-L+1)}$ and $Y_f\in\R^{pN \times (T-L+1)}$ are the portions used for prediction (i.e., future). A necessary and sufficient condition to establish that the Hankel matrix is sufficiently rich is to choose $T$ such that $T\geq (m+1)(T_\ini+N+n)-1$, which is a well-known result in behavioral systems theory.


\section{Distributed DDPC for Trajectory Planning}
\label{TRAJECTORY PLANNER}
\vspace{-0.3em}

In this section, we present our main contribution---the development of a DDPC for constrained multi-agent systems. 

\subsection{Data-Driven Predictive Control}
\label{PredictiveControl}
\vspace{-0.3em}

The aim of this section is to outline how the data-driven approach is used for predictive control, as well as highlight some difficulties. To begin, consider the real-time DeePC methodology provided in \cite{ShallowsDeePC,PowerSysDeePC} as follows:
\begin{alignat}{1}\label{DeePC}
    \min_{(u,y,g,\sigma)} \,\, & \sum^{N-1}_{k=0}\Big(\|y_{k}-y^{\des}_{k}\|^{2}_{Q}+\|u_k\|^{2}_{R}\Big)+\lambda_{g}\|g\|^2+\lambda_{\sigma}\|\sigma\|^2 \nonumber\\
    \textrm{s.t.}\,\,\,& \begin{bmatrix} U_p \\ Y_p \\ U_f \\ Y_f \end{bmatrix}g + \begin{bmatrix} 0 \\ \sigma \\ 0 \\ 0 \end{bmatrix} = \begin{bmatrix} u_\ini \\ y_\ini \\ u \\ y \end{bmatrix} \nonumber\\
    & \,u_{k}\in\pazocal{U},\quad y_{k}\in\pazocal{Y},\quad k=0,\hdots,N-1,
\end{alignat}
where $Q\in\Real^{p\times p}$ and $R\in\Real^{m\times m}$ are positive definite weighting matrices, $\|y\|^{2}_{Q}:=y^{\top}Q\,y$, $\{y^{\des}_k\}_{k=0}^{N-1}$ represents a desired trajectory, and $\pazocal{U}$ and $\pazocal{Y}$ are the feasible input and output sets, respectively. In our notation, $(u_{\ini},y_{\ini})$ denotes the past measured trajectory over the estimation horizon $T_{\ini}$, which provides feedback directly into the model. In addition, $(u,y)$ represents the predicted I-O trajectory over the control horizon $N$. This method has proven to be robust and has worked for several nonlinear systems \cite{ShallowsDeePC,PowerSysDeePC}. One of the primary reasons for this is the addition of $\lambda_{g}$ and $\lambda_{\sigma}$, which are positive weighting factors meant to regularize the $g$ vector from Theorem \ref{BST_Theorem} and penalize the defect variable $\sigma$, respectively. Note that $\sigma$ is added to lessen the effect that noisy data has on the system. If the data were to contain no noise, this variable could be removed, though it is generally required in practice. 

Although this methodology has the benefit of not requiring a direct model, it also comes with considerable computational complexity as the system increases in size due primarily to the vector $g$. Consequently, this method is intractable for real-time computation on teams of quadrupedal robots. For a more in-depth discussion on the matter, we refer the interested reader to \cite{Fawcett_Hamed_DataDriven}. To circumvent the problem, we adopt an offline approximation for $g$, analogous to \cite{PowerSysDeePC,Fawcett_Hamed_DataDriven} as follows:
\begin{alignat}{1}
    g = \begin{bmatrix}
        U_p \\ Y_p \\ U_f 
    \end{bmatrix}^\dagger \begin{bmatrix} u_\ini \\ y_\ini \\ u \end{bmatrix},\,\,
    y = \pazocal{G} \begin{bmatrix}
        u_\ini \\ y_\ini \\ u
    \end{bmatrix},\,\,
        \pazocal{G} :&= Y_f \begin{bmatrix}
        U_p \\ Y_p \\ U_f 
    \end{bmatrix}^\dagger,\label{Template}
\end{alignat}
where $(\cdot)^\dagger$ represents the pseudo inverse and $\pazocal{G}$ denotes the \textit{data-driven state transition matrix over $N$-steps}. Using this procedure, we can remove $g$ from the optimization problem \eqref{DeePC}, considerably reducing the number of decision variables.  

\subsection{Distributed Multi-Agent Trajectory Planning}
\vspace{-0.3em}

Here, we outline how the data-driven model \eqref{Template} can be used to create a distributed planner for groups of holonomically constrained legged robots. First, we present the control law in a centralized manner, i.e., assuming that one planner is used to control the whole system. This is later decomposed for distributed computation. In particular, consider the centralized predictive control problem utilizing \eqref{Template} as follows:
\begin{alignat}{1}\label{LNPC}
    \min_{(u,y)}\quad & \sum^{N-1}_{k=0} \Big(\|y_{k}-y_{k}^{\des}\|^2_{Q}+\|u_{k}\|^2_{R}\Big) \nonumber\\
    \textrm{s.t.}\quad & y = \pazocal{G} \begin{bmatrix}
        u_\ini \\ y_\ini \\ u
    \end{bmatrix} \nonumber\\
    & \,u_{k}\in\pazocal{U},\,\,y_{k}\in\pazocal{Y}, \quad \,\,k=0,\hdots,N-1.
\end{alignat}
This method has been shown to be amenable to trajectory planning for single-agent legged robots \cite{Fawcett_Hamed_DataDriven} but has not been used for multi-agent systems. In the multi-agent context, the state transition matrix $\pazocal{G}$ is created using data from all of the agents. In particular, Theorem \ref{BST_Theorem} considers the trajectory $(u^{\textrm{d}},y^{\textrm{d}})$ to define the Hankel matrix. To construct the Hankel martix for multi-agent systems, we can define $u^{\textrm{d}}:=\{\col(u^{\textrm{d},1}_k,u^{\textrm{d},2}_k,\cdots,u^{\textrm{d},n_{a}}_k)\}_{k=0}^{T-1}$ and $y^{\textrm{d}}:=\{\col(y^{\textrm{d},1}_k,y^{\textrm{d},2}_k,\cdots,y^{\textrm{d},n_{a}}_k)\}_{k=0}^{T-1}$, where $(\cdot)^{\textrm{d},i}_{k}$ for all $i\in\pazocal{I}:=\{1,\cdots,n_{a}\}$ denotes the data contributed by agent $i$ at the sample time $k$, and $n_{a}$ is the total number of agents. Using these combined I-O pairs, we obtain a large Hankel matrix describing the entire complex system. Furthermore, this new Hankel matrix can be decomposed according to \eqref{HankelDecomp}, and the corresponding $g$ vector can be approximated using \eqref{Template}.

In moving to multi-agent systems, however, it is desirable to share the computational load between agents. In order to do so, we consider a decomposition of $\pazocal{G}$ as follows:
\begin{align}
    \pazocal{G} = \begin{bmatrix}
        \pazocal{G}_{1,1} & \pazocal{G}_{1,2} & \cdots & \pazocal{G}_{1,n_{a}} \\
        \pazocal{G}_{2,1} & \pazocal{G}_{2,2} & \cdots & \pazocal{G}_{2,n_{a}} \\
        \vdots & \vdots & \ddots & \vdots \\
        \pazocal{G}_{n_{a},1} & \pazocal{G}_{n_{a},2} & \cdots & \pazocal{G}_{n_{a},n_{a}}
    \end{bmatrix},
\end{align}
where $\pazocal{G}_{i, j}$ represents the effect of agent $j$ on the predicted output of agent $i$ (i.e., coupling). In particular, the predicted output of agent $i\in\pazocal{I}$ can be expressed by
\begin{equation}\label{predicted_output}
    y^{i} = \pazocal{G}_{i,i} \begin{bmatrix} u^{i}_\ini \\ y^{i}_\ini \\ u^{i} \end{bmatrix}
    + \sum_{j\neq i} \pazocal{G}_{i,j} \begin{bmatrix} u^{j}_\ini \\ y^{j}_\ini \\ u^{j} \end{bmatrix},
\end{equation}
where $(\cdot)^{i}$ represents the variables related to the $i$th agent. To alleviate the coupling problem, we adopt a \textit{one-step communication delay protocol}. In particular, we assume that at every time sample $t$, each local DDPC has access to the optimal predicted solutions of the other local DDPCs and their past measurements at time $t-1$. Using this assumption, the predicted output of agent $i$ can be approximated by
\begin{equation}
    y^{i} \approx \pazocal{G}_{i,i}\begin{bmatrix} u^{i}_\ini \\ y^{i}_\ini \\ u^{i} \end{bmatrix} + \sum_{j\neq i} \pazocal{G}_{i,j} \begin{bmatrix} u^{j}_{\ini |t-1} \\ y^{j}_{\ini |t-1} \\ u^{j}_{|t-1} \end{bmatrix},
\end{equation}
where $u^{j}_{|t-1}$ denotes the optimal solution of the local DDPC for agent $j$ at time $t-1$, wherein the summation is constant until the next time the DDPC is run. In addition, $(u^{j}_{\ini |t-1},y^{j}_{\ini |t-1})$ represents the past measurements of agent $j$ at time $t-1$. This choice allows each local planner to run independently and results in the following network of distributed DDPCs (local QPs) to be solved at time $t$ 
\begin{alignat}{1}\label{Distributed}
    \min_{(u^{i},y^{i})}\quad & \sum^{N-1}_{k=0} \Big(\|y^{i}_{k}-y_{k}^{\des,i}\|^2_{Q}+\|u^{i}_{k}-u_{k}^{\des,i}\|^2_{R}\Big) \nonumber\\
    \textrm{s.t.}\quad & y^{i} = \pazocal{G}_{i,i}\begin{bmatrix} u^{i}_\ini \\ y^{i}_\ini \\ u^{i} \end{bmatrix} + \sum_{j\neq i} \pazocal{G}_{i,j} \begin{bmatrix} u^{j}_{\ini |t-1} \\ y^{j}_{\ini |t-1} \\ u^{j}_{|t-1} \end{bmatrix} \nonumber\\
    &u^{i}_{k}\in\pazocal{U},\,\,y^{i}_{k}\in\pazocal{Y}, \quad k= 0,\hdots,N-1,
\end{alignat}
where $u_{k}^{\des,i}$ and $y_{k}^{\des,i}$ represent the desired inputs and outputs for agent $i$ at the prediction step $k$ (see Fig. \ref{OverviewFig}). The optimal input and output trajectories are then passed to the low-level controller for tracking. In this work, we choose a subset of COM state variables for the output $y$ while taking the ground reaction forces (GRFs) as the control inputs $u$. This will be clarified more in Section \ref{DATA_COLLECTION}. Consequently, the feasible set $\pazocal{U}$ is chosen as the linearized friction cone, i.e.,  $\pazocal{U}=\pazocal{FC}:=\{\col(f_{x},f_{y},f_{z})|f_{z}>0, \, |f_{x}|\leq\frac{\mu}{\sqrt{2}}f_{z}, \,|f_{y}|\leq\frac{\mu}{\sqrt{2}}f_{z}\}$, where $\mu$ denotes the friction coefficient. 


\begin{figure*}[t!]
\centering
\includegraphics[width=\linewidth]{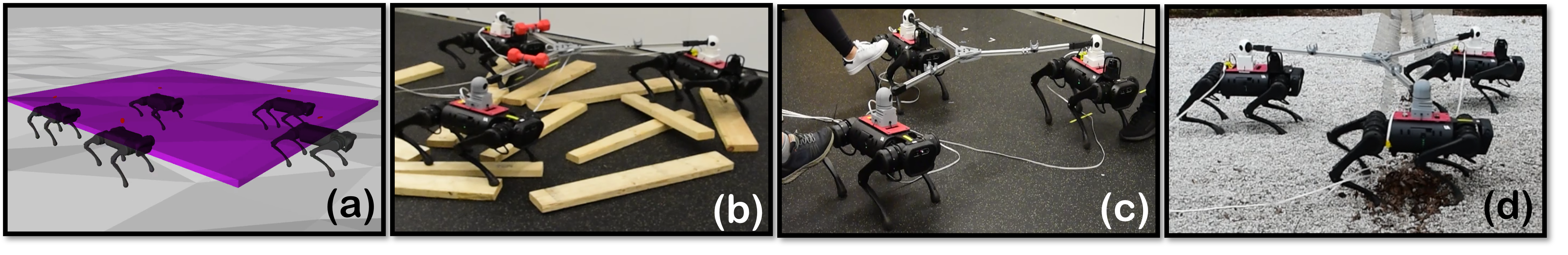}
\vspace{-1.8em}
\caption{(a) Simulation results of $5$ agents over varying terrain with a payload of $10$ (kg), (b) rough terrain experiment with unstructured wooden blocks and a $4.5$ (kg) payload, (c) experiment with push disturbances, and (d) experiment maneuvering over gravel. Videos are available online \cite{YouTube_Link}.}
\label{Snapshots}
\vspace{-1.2em}
\end{figure*}

\section{Nonlinear Low-Level Controller}
\label{LOWLEVEL}
\vspace{-0.3em}

The purpose of this section is to briefly provide the details of the low-level controller \cite{Fawcett_Hamed_CLF}, based on QP and virtual constraints \cite{Jessy_Book}, to be used in this work. In particular, we model each robot using a floating base and represent the generalized coordinates of the system by $q\in\pazocal{Q}\subset\Real^{n_{q}}$, where $\pazocal{Q}$ is the configuration space and $n_{q}$ represents the number of degrees of freedom (DOFs) of the system. We further denote the joint-level torques by $\tau\in\pazocal{T}\subset\Real^{m_{\tau}}$. In this notation, $\pazocal{T}$ denotes the allowable torques and $m_{\tau}$ denotes the number of actuators. It is assumed that the interaction wrenches between agents are encapsulated by the data-driven model. Therefore, we neglect these wrenches in the low-level controller. In particular, the overarching equations of motion for the full-order system of each agent become
\begin{equation}\label{Dynamics}
    M(q)\,\ddot{q}+H(q,\dot{q})=\Upsilon\,\tau + J^\top(q)\,f,
\end{equation}
where $M(q)\in\Real^{n_{q}\times n_{q}}$ is the mass-inertia matrix, $H(q,\dot{q})\in\Real^{n_{q}}$ denotes the Coriolis, centrifugal, and gravitational terms, $\Upsilon\in\Real^{n_{q}\times{}m_{\tau}}$ represents the input matrix, $J(q)$ denotes the contact Jacobian matrix, and $f$ represents the GRFs at the stance feet. We further suppose that the positions of the stance leg ends, denoted by $r$, do not slip, i.e., $\ddot{r}=0$. 

With the dynamics in hand, we can now present the virtual constraints controller. In particular, we aim to track both force and COM trajectories generated by the high-level distributed planners. For this purpose, we consider virtual constraints as output functions to be regulated as $h(q,t):=h_{0}(q)-h^{\des}(t)$. These virtual constraints are then imposed via partial feedback linearization \cite{Isidori_Book}. Here, $h_{0}(q)$ denotes the controlled variables consisting of the COM position, orientation, and the Cartesian coordinates of the swing leg ends. Finally, $h^{\des}(t)$ represents the desired evolution of $h_{0}(q)$. In this work, the desired end position for a swing leg is chosen using the Raibert heuristic \cite[Eq. (4), pp. 46]{raibert1986legged}, and the trajectory for the swing leg is defined using a B\'ezier polynomial. These virtual constraints, along with the no slippage condition, are concatenated into a single strictly convex QP to be solved at 1kHz as follows \cite{Fawcett_Hamed_CLF}
\begin{alignat}{4}\label{QP}
    &\min_{(\tau,f,\delta)} \,\,\,&& \frac{\gamma_{1}}{2}\|\tau\|^2 + \frac{\gamma_{2}}{2}\|f - f^{\des}\|^2 + \frac{\gamma_{3}}{2} && \|\delta\|^2 \nonumber\\
    &\textrm{s.t.} && \ddot{h}(\tau,f) = -K_{P}\,h - K_{D}\,\dot{h} + \delta && \textrm{(Output Dynamics)}\nonumber\\
    & && \ddot{r}(\tau,f) = 0 \nonumber && \textrm{(No slippage)}\\
    & && \tau\in\pazocal{T},\quad f\in\pazocal{FC} && \textrm{(Feasibility)},
\end{alignat}
where $\gamma_{1}$, $\gamma_{2}$, and $\gamma_{3}$ are positive weighting factors, and the desired force profile $f^{\des}(t)$ (i.e., inputs $u$) is prescribed by the high-level DDPC in \eqref{Distributed}. The equality constraints are expressed as 1) the output dynamics $\ddot{h} + K_{D}\,\dot{h} + K_{P}\,h = \delta$ for positive definite gain matrices $K_{P}$ and $K_{D}$ and $\delta$ being a defect variable to ensure feasibility, and 2) the no slippage condition $\ddot{r}=0$. We remark that $\ddot{h}$ and $\ddot{r}$ are affine functions of $(\tau,f)$, hence, the problem is convex.  We direct the reader to \cite[Appendix A]{kim2022cooperative} for more information regarding the derivation of $\ddot{h}$ and $\ddot{r}$ according to Lie derivatives. The QP solves for the minimum-power torques $\tau$ while tracking the prescribed forces and COM trajectory.




\section{Experiments}
\vspace{-0.4em}

In this section, we provide the procedure for collecting the data for the reduced-order model, and further, provide the simulation and experimental results. Here we consider the $18$-DOF quadruped A1 made by Unitree. The robot is modeled using a floating base, with the first $6$ DOFs being composed of the unactuated position and orientation of the trunk. The remaining DOFs are composed of the actuated hip roll, hip pitch, and knee pitch joints for each leg. The robot weighs around $12.45$ (kg) and the center of the trunk is $0.26$ (m) above the ground during locomotion. We are, however, interested in multi-agent systems. We assume that each of the agents is holonomically constrained to other agents in a complete graph, i.e., there is no relative translational motion between agents. In order to accomplish this, we connect the robots together rigidly through a ball joint (see Fig. \ref{Snapshots}). In simulations, this is imposed via a distance constraint.


\subsection{Data Collection}\label{DATA_COLLECTION}
\vspace{-0.3em}

We begin by describing the I-O pairs considered in this work and the manner in which the data was collected. The data that is to be used in the Hankel matrix is collected in simulation at $100$ (Hz) during nominal locomotion. This is in contrast to \cite{Fawcett_Hamed_DataDriven}, which considered data collected on hardware. However, one of the contributions of this work is to create a reduced-order model of highly complex systems that interact with one another. From this standpoint, it makes sense to consider simulation data since each individual agent can be modeled accurately, but the complexity of collaborating agents is prohibitive in terms of defining a physics-based reduced-order model. This also shows good sim-to-real transfer for the learned model, as will be shown in what follows.

In order to perform the data collection in simulation, we first choose the inputs to be the forces at the contacting leg ends and and we take the outputs as $y^{\textrm{d}}=\col(\textrm{z},\dot{\textrm{x}},\dot{\textrm{y}},\textrm{roll},\textrm{pitch},\omega_{z})$, where $\textrm{z}$ is the standing height, $\dot{\textrm{x}}$ and $\dot{\textrm{y}}$ are the linear velocities of the COM in the transverse plane, and $\omega_z$ denotes the angular velocity about the vertical axis of the torso. These outputs are chosen because they represent the variables of interest to an end-user when providing joystick commands to a robot. It should be noted that other I-O realization could provide fruitful results depending on the goal of the planner and the available measurements. During the data collection procedure, the quadrupeds are commanded to walk around in RaiSim \cite{RAISIM} using just the low-level controller \eqref{QP}, while random noise is injected into the desired forces, which helps ensure the persistence of excitation. In particular, we choose the desired forces to be $u^{\textrm{des},i}_{k,\ell}:=\col(0,0,\frac{m^{\textrm{net},i}g_0}{N^{i}_{c,k}})$ for each contacting leg $\ell\in\pazocal{C}_{k}^{i}$ and zero otherwise, where $m^{\textrm{net},i}$ is the total mass of agent $i$, $g_0$ is the gravitational acceleration, $N_{c,k}^{i}$ is the anticipated number of contacting legs for agent $i$ at time $k$, and $\pazocal{C}_{k}^{i}$ is the anticipated set of contacting legs for agent $i$ at time $k$. That is, when collecting data, we choose the desired force to be a random perturbation about the nominal amount of force that is required to hold the quadruped in a static position based on the number of anticipated contacts. 

For this problem, we consider an estimation horizon of $T_{\ini}=10$ and $N=25$ for the prediction horizon. We further ensure that $T$ is chosen such that the amount of data collected far exceeds that required by the general theory. This, in turn, assists in potentially capturing more nonlinear information while it also reduces the impact of noise. The collection of additional data when utilizing the formulation \eqref{DeePC} could pose an issue due to an increase in decision variables. However, this is mitigated by utilizing the approximation \eqref{Template}. 

\begin{figure}[t!]
\centering
\includegraphics[width=\linewidth]{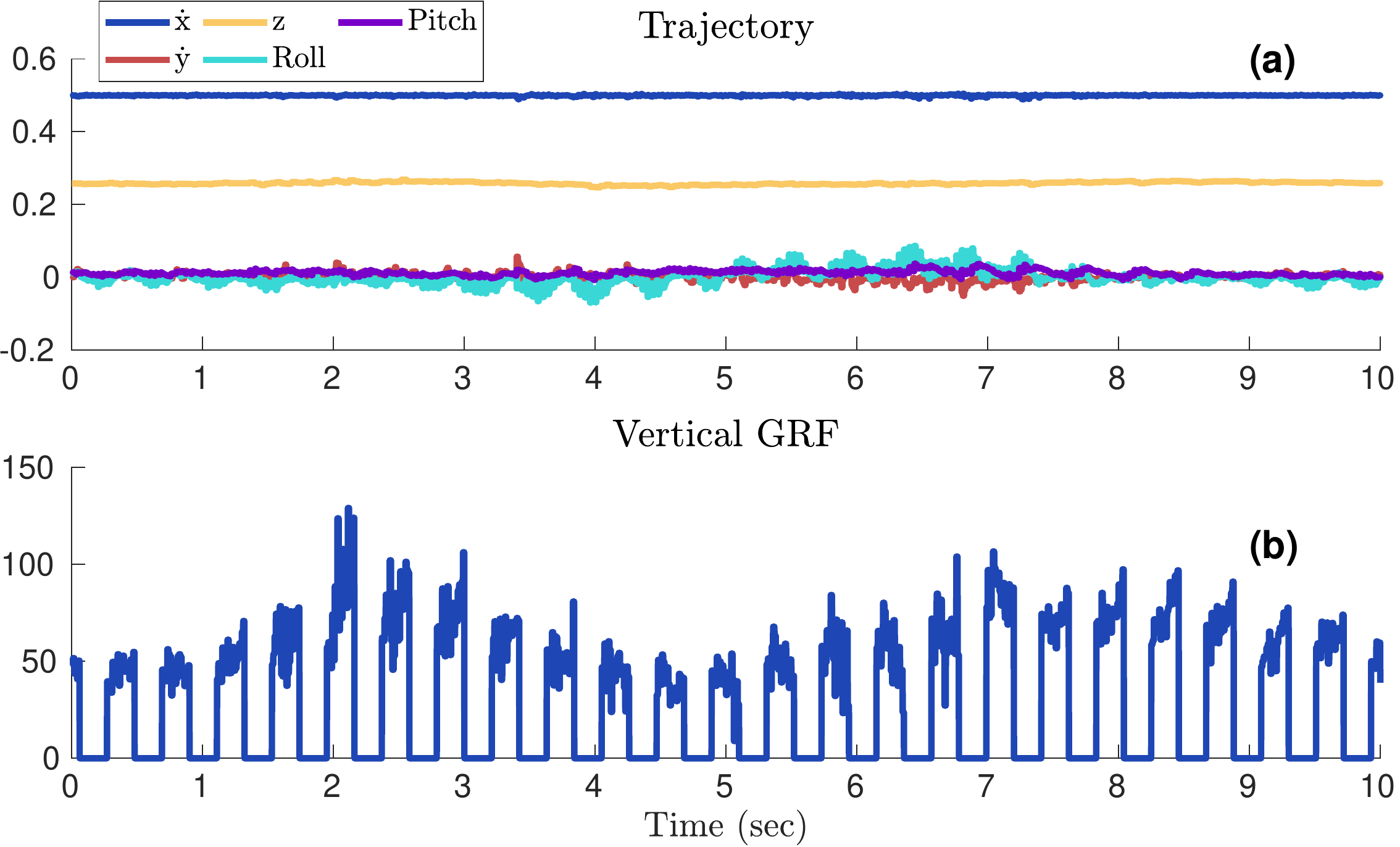}
\vspace{-2.0em}
\caption{The trajectory from the planner (a) and the vertical GRF of the front right leg from the planner (b) for agent $1$. A forward speed of $0.5$ (m/s) is commanded, and the standing height is $0.26$ (m). The multi-agent system is subject to uneven terrain and a payload of $10$ (kg), and can maneuver robustly. A snapshot of the simulation can be found in Fig. \ref{Snapshots} (a).}
\label{SimPlot}
\vspace{-1.0em}
\end{figure}

\subsection{Simulation Experiments}
\vspace{-0.3em}

The high-level planner contains $450$ decision variables per agent. In the centralized case using $5$ agents shown here, that amounts to $2250$ decision variables, which further motivates the necessity for a distributed approach. Under the distributed scheme, the predictive controller is updated every $40$ (ms) ($25$ (Hz)), and the first $4$ time steps are implemented, i.e., we predict over a horizon of $250$ (ms) and implement the first $40$ (ms) of the prediction. The high-level planner is solved using OSQP \cite{osqp} and takes approximately $15$ (ms) on an external laptop with an Intel\textsuperscript{\textregistered} Core\textsuperscript{\texttrademark} i7-1185G7 running at $3.00$ GHz and $16$ GB of RAM. However, solve times of $\sim30$ (ms) have been observed, further motivating the decision to update the planner at $40$ (ms). The predictive controller parameters are chosen as $Q=\textrm{diag}(1e6,1e5,1e5,2e5,1e5,1e4)$ and $R=I\otimes\textrm{diag}(0.05,0.05,0.5)$, where $I$ is the identity of appropriate size, and $\otimes$ represents the Kronecker product. Finally, the parameters used by the low-level controller to track the trajectory and forces from the planner are chosen to be $\gamma_{1}=10^{2}$, $\gamma_{2}=10^{3}$, and $\gamma_{3}=10^{6}$, resulting in stable locomotion. All of the gains used in simulation are the same as those used on hardware.


In order to show the efficacy of the proposed controller for $5$ agents, we consider a compound experiment in simulation such that the multi-agent system is subject to an unknown payload of $10$ (kg) and uneven terrain. A snapshot of the simulation can be found in Fig. \ref{Snapshots} (a), while the prescribed forces and trajectory for the first agent can be found in Fig. \ref{SimPlot}. From these figures, it is evident that the planner produces forces that are feasible, while also resulting in a viable COM trajectory for the low-level controller to track. 

\begin{figure}[t!]
\centering
\includegraphics[width=\linewidth]{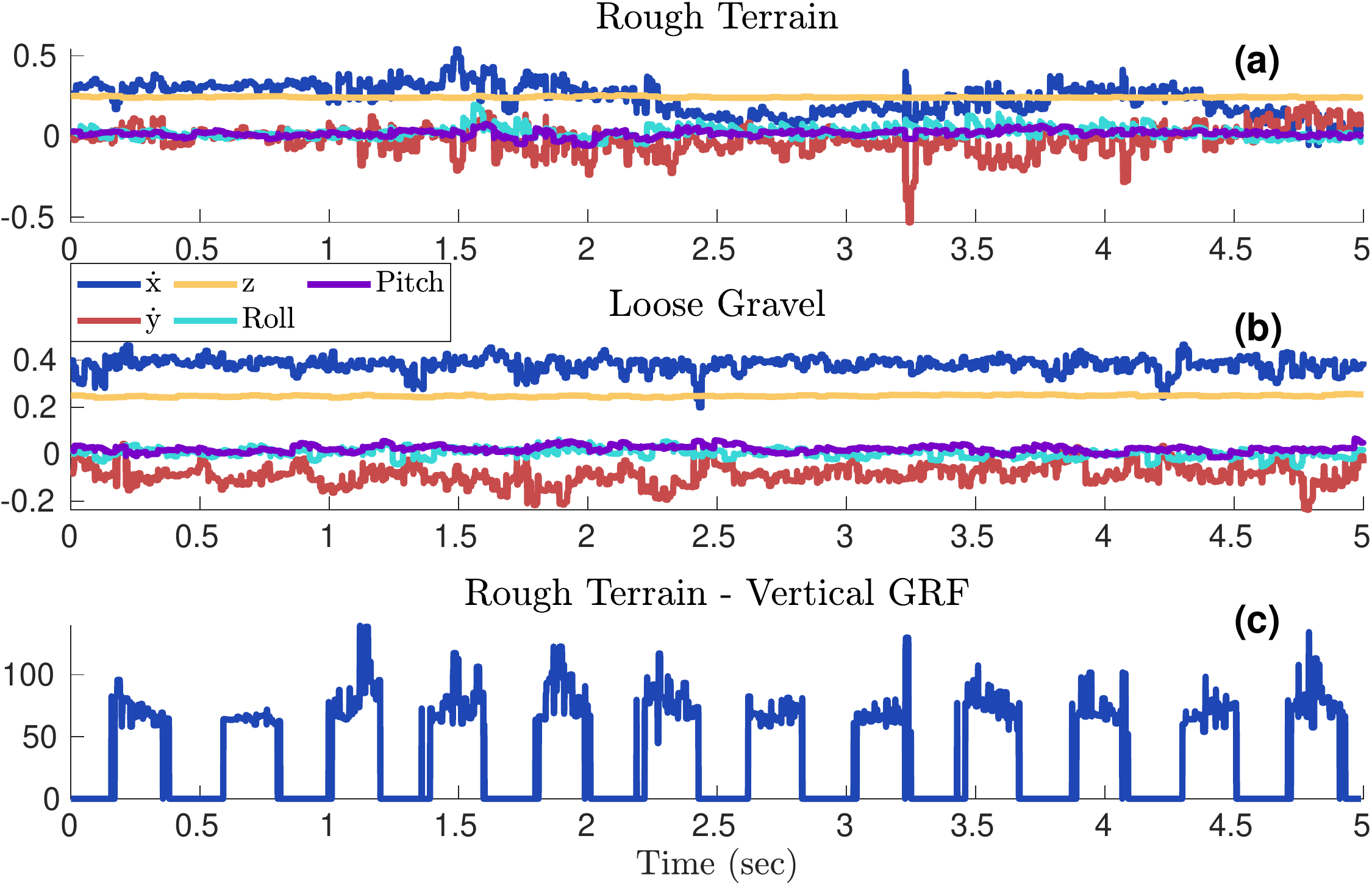}
\vspace{-2.0em}
\caption{The trajectory from the planner of agent $1$ while trotting at approximately $0.4$ (m/s) subject to (a) rough terrain with unstructured wooden blocks and (b) loose gravel. We further show the vertical GRF for the front right leg produced by the planner for rough terrain in (c). The GRF during the gravel experiment is similar.}
\label{HL_traj_and_force}
\vspace{-1.em}
\end{figure}

\begin{figure}[t!]
\centering
\includegraphics[width=\linewidth]{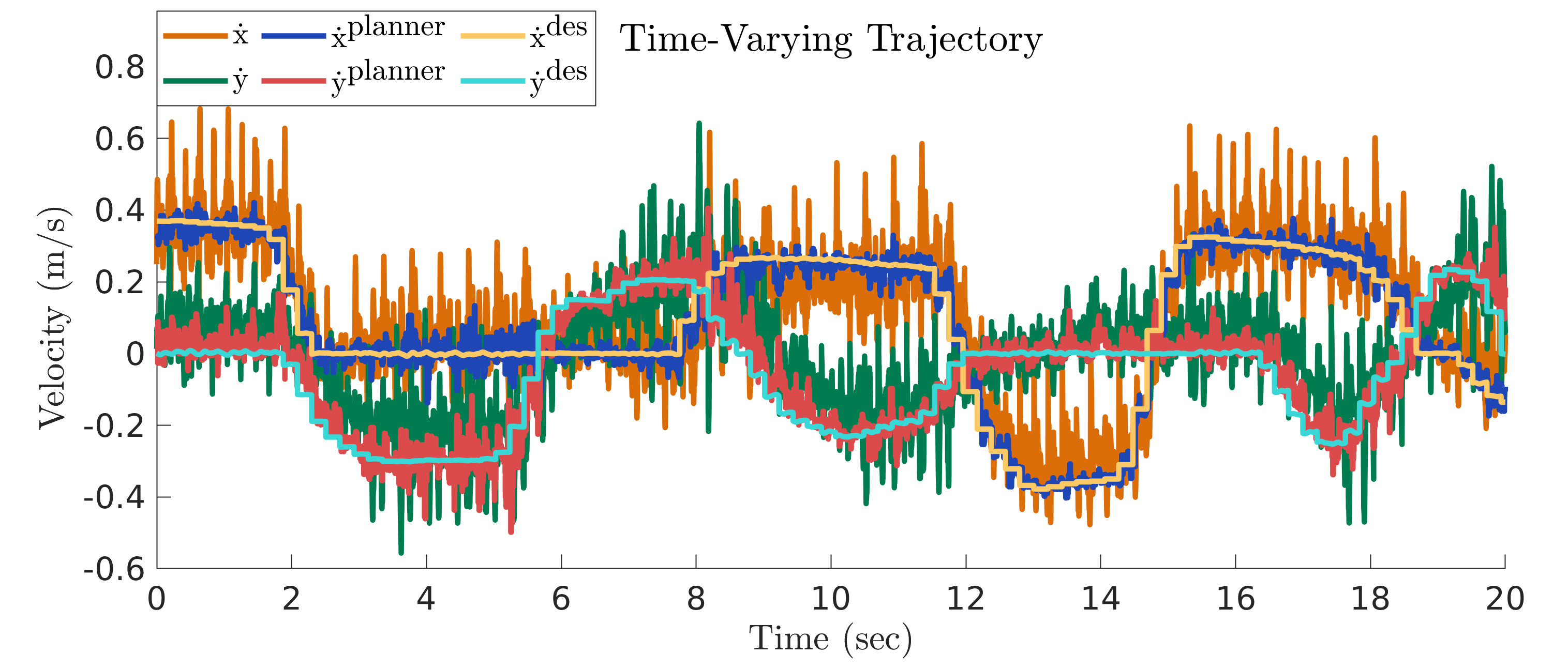}
\vspace{-2.0em}
\caption{The plot shows the tracking performance of the planner when using a time-varying trajectory provided using a joystick. In this experiment, the quadrupeds navigate flat ground subject to a $6.8$ (kg) payload. The plot shows the trajectory of agent $1$.}
\label{HL_Joystick}
\vspace{-1.2em}
\end{figure}

\subsection{Hardware Experiments}
\vspace{-0.3em}

Finally, we provide hardware experiments to show the effectiveness of the planner. In particular, for hardware experiments, we consider the use of $3$ quadrupeds that are holonomically constrained using a ball joint. Snapshots of several experiments can be found in Fig. \ref{Snapshots} (b)-(d), which shows the multi-agent system subject to external disturbances and unknown environments. In Fig. \ref{HL_traj_and_force}, we illustrate the behavior of the planner in terms of its prescribed trajectory when walking over unstructured wooden blocks (Fig. \ref{HL_traj_and_force} (a)) and navigating over loose gravel (Fig. \ref{HL_traj_and_force} (b)). The corresponding forces for the rough terrain experiment can be found in Fig. \ref{HL_traj_and_force} (c). Finally, we provide an additional experiment to show the ability of the planner to track a time-varying trajectory subject to a $6.8$ (kg) payload, where the trajectory produced by the planner can be found in Fig. \ref{HL_Joystick}. It is evident that the planner provides a robustly stable output even in the presence of significant uncertainty in terms of payloads and various environmental factors. Videos of the simulation and hardware experiments can be found in \cite{YouTube_Link}.


\section{Conclusion}
\vspace{-0.4em}
This work presented a data-driven planner for robust multi-agent quadrupedal locomotion, wherein the robots were constrained to one another with a ball joint. We considered the use of behavioral systems theory to model the complex system and further proposed a distributed scheme to spread the computational load. We provided extensive experiments both in simulation and on hardware, which showed the robustness of this method to uncertainty in terrain, payloads, and external disturbances. Future work will examine this methodology when the robots do not form a complete graph, i.e., the agents are constrained but have limited freedom to change formation. Additionally, we will explore how this method could extend to an even greater number of agents. 

\vspace{-0.3em}
\bibliographystyle{IEEEtran}
\balance
\bibliography{references}

\end{document}